\documentclass[sigconf]{acmart}
\AtBeginDocument{%
  }

\copyrightyear{2025}
\acmYear{2025}
\setcopyright{acmlicensed}
\acmConference[WWW '25] {Proceedings of the ACM Web Conference 2025}{April 28-May 2, 2025}{Sydney, NSW, Australia.}
\acmBooktitle{Proceedings of the ACM Web Conference 2025 (WWW '25), April 28-May 2, 2025, Sydney, NSW, Australia}
\acmISBN{979-8-4007-1274-6/25/04}
\acmDOI{10.1145/3696410.3714520}
  
\usepackage{algorithm}
\usepackage{algpseudocode}
\usepackage{booktabs}
\usepackage{amsmath}
\usepackage{graphicx} 
\usepackage{multirow}
\usepackage{array} 
\usepackage{tabularx} 
\usepackage{enumerate}

\begin{document}

\title{Grad: Guided Relation Diffusion Generation for Graph Augmentation in Graph Fraud Detection}

\author{Jie Yang}
\authornote{Both authors contributed equally to this research.}
\affiliation{%
  \institution{Tongji University}
  \city{Shanghai}
  \country{China}
  }
\email{2153814@tongji.edu.cn}

\author{Rui Zhang}
\authornotemark[1]
\affiliation{%
  \institution{The University of New South Wales}
  \city{Sydney}
  \country{Australia}
  }
\email{royzhz@outlook.com}

\author{Ziyang Cheng}
\affiliation{%
  \institution{Tongji University}
  \city{Shanghai}
  \country{China}
  }
\email{2151393@tongji.edu.cn}

\author{Dawei Cheng}
\authornote{Corresponding author.}
\affiliation{%
  \institution{Tongji University \& Shanghai Artificial Intelligence Laboratory}
  \city{Shanghai}
  \country{China}
  }
\email{dcheng@tongji.edu.cn}

\author{Guang Yang}
\affiliation{%
  \institution{Wechat Pay, Tencent Inc.}
  \city{Shenzhen}
  \country{China}
  }
\email{mecoolyang@tencent.com}

\author{Bo Wang}
\affiliation{%
  \institution{Wechat Pay, Tencent Inc.}
  \city{Shenzhen}
  \country{China}
  }
\email{pollowang@tencent.com}

\renewcommand{\shortauthors}{Jie Yang et al.}

\begin{abstract}

Nowadays, Graph Fraud Detection~(GFD) in financial scenarios has become an urgent research topic to protect online payment security.
However, as organized crime groups are becoming more professional in real-world scenarios, fraudsters are employing more sophisticated camouflage strategies.
Specifically, fraudsters disguise themselves by mimicking the behavioral data collected by platforms, ensuring that their key characteristics are consistent with those of benign users to a high degree, which we call Adaptive Camouflage.
Consequently, this narrows the differences in behavioral traits between them and benign users within the platform's database, thereby making current GFD models lose efficiency.
To address this problem, we propose a relation diffusion-based graph augmentation model Grad. 
In detail, Grad leverages a supervised graph contrastive learning module to enhance the fraud-benign difference and employs a guided relation diffusion generator to generate auxiliary homophilic relations from scratch. 
Based on these, weak fraudulent signals would be enhanced during the aggregation process, thus being obvious enough to be captured.
Extensive experiments have been conducted on two real-world datasets provided by WeChat Pay, one of the largest online payment platforms with billions of users, and three public datasets. The results show that our proposed model Grad outperforms SOTA methods in both various scenarios, achieving at most 11.10\% and 43.95\% increases in AUC and AP, respectively. 
Our code is released at \url{https://github.com/AI4Risk/antifraud} and \url{https://github.com/Muyiiiii/WWW25-Grad}.

\vspace{-2mm}

\end{abstract}



\received{20 February 2007}
\received[revised]{12 March 2009}
\received[accepted]{5 June 2009}

\maketitle

\section{Introduction}

\begin{figure}[t]
  \centering
  \includegraphics[width=\linewidth]{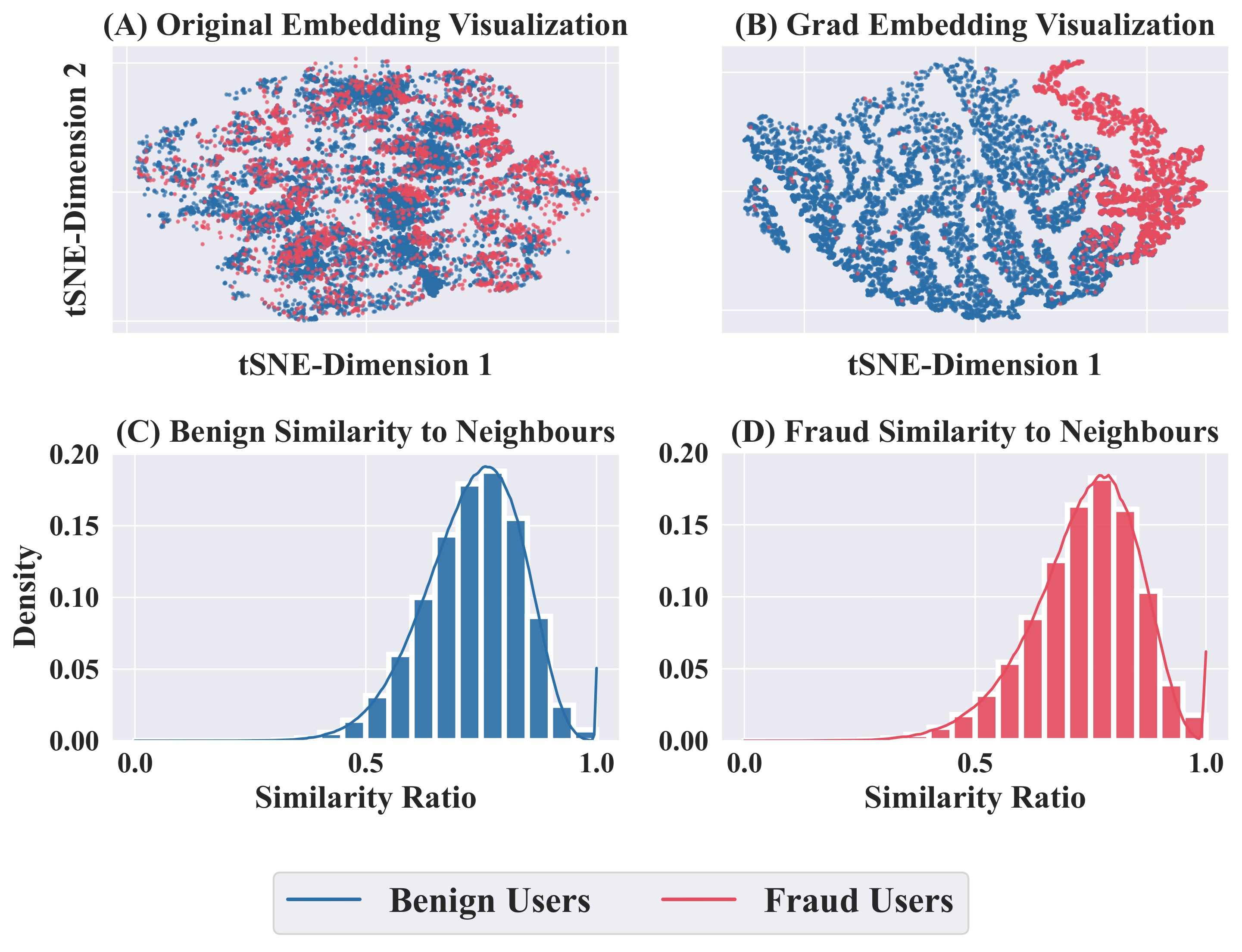}
  \caption{ The analysis of features in benign nodes and fraudulent nodes with Adaptive Camouflage. 
  (A, B): the visualization of the feature embeddings of original and after-Grad nodes. (C, D): the similarity distribution between benign and fraudulent nodes and their neighbors.}
  
  \label{fig:dataAnalysis}
\end{figure}

With the rapid development of computer information technology, the increasing prevalence of fraudulent activities \cite{deng2021graph,cheng2023anti, cheng2022regulating} have been posing significant threats to the development of online financial platforms. 
According to the latest report from the Federal Trade Commission~(FTC), fraud losses increased more than 30\% over the last year, reaching nearly \$8.8 billion \cite{commission2023consumer}. 
Consequently, financial fraud detection is becoming an increasingly important topic in protecting online payment security and building the trustworthiness of online financial systems \cite{khedmati2020applying, gao2023rumor, ma2023fighting}.

Due to the inherent spatial property of financial-related networks, where users can be represented as nodes and transactions can be viewed as edges of graphs naturally, recent researchers have utilized graphs-based methods to identify and characterize fraud behaviors \cite{cheng2021causal, zhang2024pre}. 
Among them, Graph Neural Networks~(GNNs) \cite{liu2024towards, ma2023towards, roy2024gad} show superior capability in processing the graph data that is non-Euclidean with diverse graph structures and node attributes. 
As a result, GNNs have been employed for financial fraud detection and achieved remarkable success \cite{zhang2022dual, yanqiao2020deep, zhuang2023subgraph}.

Unfortunately, as organized crime groups are becoming more professional in real-world scenarios \cite{qiao2024truncated, wu2024safeguarding, cheng2018modeling}, fraudsters are employing more sophisticated camouflage strategies tailored to platforms' detection methods. 
Specifically, fraudsters disguise themselves by mimicking the behavioral data collected by platforms, ensuring that their key characteristics are consistent with those of benign users~\cite{cheng2025graph, han2025mitigating}. 
This reduces the differences in behavioral traits between them and benign users within the platform's database, thereby disturbing detectors' judgment. 
Generally, in the real world, online payment platforms typically collect users' transaction histories, including details like transaction partners and amounts, to build user features. 
Fraudsters exploit this by accumulating transaction data similar to that of benign users and only engaging in minimal fraudulent activities at random times, thus making their characteristics appear normal.
In this paper, we call this high-intensity camouflage strategy as Adaptive Camouflage. 
As shown in Figure \ref{fig:dataAnalysis}, in the WeChat Pay platform, the embedding distribution and neighbor similarity of fraudsters are highly consistent with benign users.
Specifically, they have an extremely high similarity ratio~(83.2\%) to benign users, which far exceeds the fraud-benign similarity ratio~(41.8\%) of conventional fraud detection dataset Amazon through our analysis.
Consequently, how to extract the remaining faint fraudulent signals~(16.8\%) caused by Adaptive Camouflage is becoming the key to real-world fraud detection.

From the perspective of detecting faint signals of fraud, existing works have achieved progress in two aspects: some employ spectral/spatial domain methods to extract weak anomalous information from the original graph directly, while others utilize generative methods to augment weak fraudulent signals. The former methods mostly focus on the high-frequency anomalous components \cite{tang2022rethinking,
gao2023addressing} and function as adaptive filters \cite{nt2019revisiting} to identify fraud. 
Furthermore, the generative methods \cite{wu2023graph, meng2023generative, yang2024directional} aim to learn the latent distribution of data to generate new fraudulent samples, which can enhance the fraudulent signals and widen the decision boundary between benign and fraudulent nodes.

Nevertheless, the emergency of sophisticated Adaptive Camouflage poses significant challenges to these methods, because they are still stuck in the research stage and far from practical deployment in real industry.
Firstly, most spectral/spatial domain methods are primarily designed for traditional camouflage behaviors, such as those observed on Amazon Dataset, where the similarity between fraudulent and benign users is approximately 41.8\% \cite{mcauley2013amateurs, wang2019semi, cui2020deterrent}.
In cases of real-world Adaptive Camouflage~(e.g. 83.2\%), these methods tend to prioritize low-frequency benign signals. 
Consequently, they essentially function as low-pass filters and ultimately fail to detect the truly valuable, weak fraudulent signals \cite{nt2019revisiting}.
In addition, for those generative-based methods, Adaptive Camouflage casts a detrimental influence on their ability to learn fraud distribution and position new fraudulent nodes appropriately.
Such an issue would lead to significant differences between generated and existing fraudulent nodes, thus limiting their fraud detection performance.

To address the serious issues posed by Adaptive Camouflage, we propose a relation diffusion-based graph augmentation model Grad. 
Firstly, we utilize a supervised graph contrastive learning method to mine valuable expert knowledge from the limited labels so that we can enhance the decision boundary between fraudulent and benign users.
Then we employ a guided relation diffusion generator module, under the guidance of the supervised graph contrastive learning method and a degree penalty function, to generate auxiliary graph relations that are free of inter-class edges. 
Finally, we use a weighted multi-relation filters to
continuously amplify weak fraudulent signals during the aggregation process and capture the weak fraudulent information through multiple flexible, spatial/spectral-localized, band-pass filters.
The effectiveness of our proposed method has been demonstrated by the validation of three public datasets and the WeChat Pay\footnote{https://www.wechat.com/} platform—one of the largest online payment platforms, with billions of users and transactions daily.
Our main contributions can be summarized as follows:

\begin{itemize}

\item To the best of our knowledge, this is the first work to address Adaptive Camouflage with significantly high fraud-benign similarity ratios by proposing a relation diffusion-based graph augmentation model, which can amplify the weak fraudulent signals and widen the decision boundary between fraudulent and benign users.

\item We devise a supervised graph contrastive learning method to widen the decision boundary between fraudulent and benign users and a guided relation diffusion generator to produce homophilic auxiliary relations. They jointly assist the final detector in capturing the weak fraudulent signals in the real-world scenario.

\item We conduct comprehensive experiments on three public datasets and large-scale real-world datasets provided by WeChat Pay. Compared with current models, Grad achieves at most over 10\% and 40\% increases in AUC and AP, respectively, which demonstrate the effectiveness of our model. 

\end{itemize}

\begin{figure*}[htbp]
\centering
\includegraphics[width=1\textwidth]{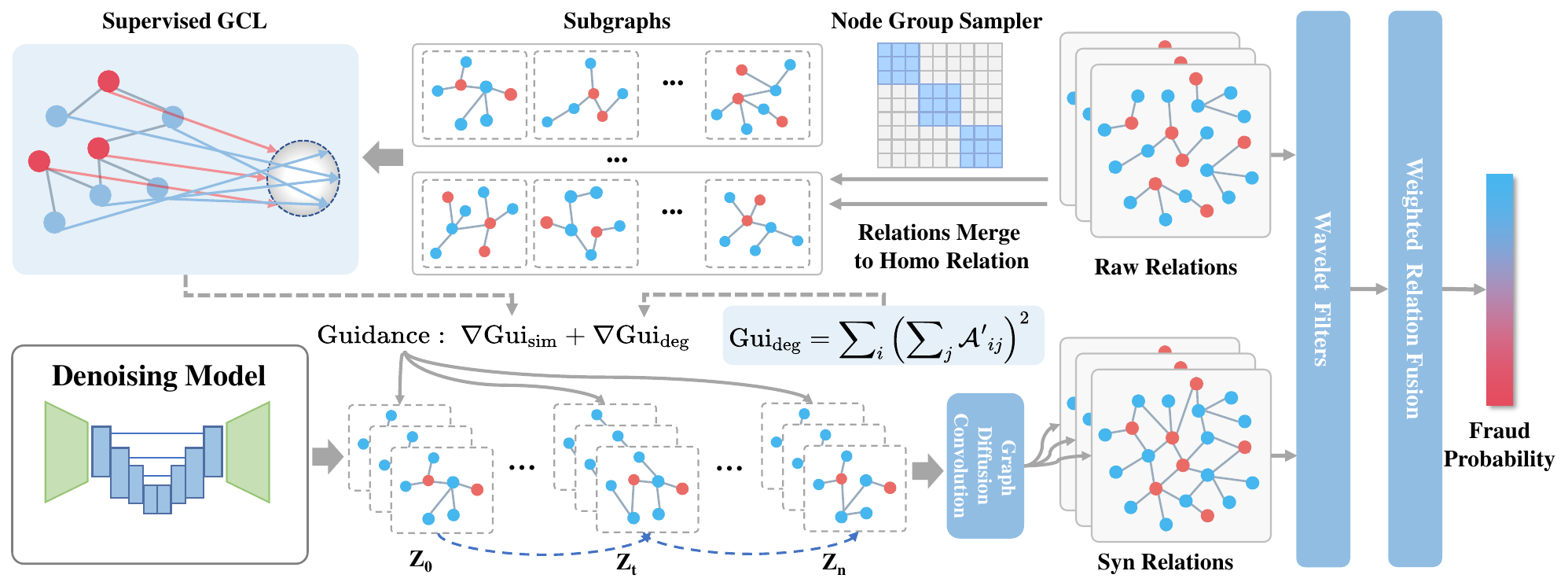}
\caption{The framework of Grad with five main components: (A) Node Group Sampler splits the entire graph into equal-sized non-overlapping subgraphs; (B) Supervised Graph Contrastive Learning~(GCL) module enhances fraud-benign difference based on valuable labels; (C) Guided Relation Diffusion Generator generates homophilic auxiliary relations from scratch; (D) Relation Augmentation module provides extra global information of the entire financial transaction networks; (E) Multi-Relation Detector fuses multiple relations and detects fraudulent signals for accurate fraud detection.}

\label{fig:model}
\end{figure*}

\section{Related Work}

\subsection{Graph Fraud Detection}

Graph fraud detection aims to detect fraudulent nodes accurately.
The methods of this task can be categorized into two types: spatial domain \cite{zhang2024generation, ghosh2023gosage, wang2019semi} and spectral domain methods \cite{tang2022rethinking, gao2023addressing}. 
Firstly, in the spatial domain, existing works employ a variety of mechanisms to distinguish homophilic and heterophilic neighbors from the node representation perspective, e.g.\ GoSage \cite{ghosh2023gosage}, SemiGNN \cite{wang2019semi} and  DETERRENT \cite{cui2020deterrent} employ attention mechanisms to give neighbors different attention scores, thereby determining the information each neighbor transmits, DGA-GNN \cite{duan2024dga}, CARE-GNN \cite{dou2020enhancing} and PC-GNN \cite{liu2021pick} utilize various sampling strategies to selectively aggregate neighborhood information, and SAD \cite{tian2023sad} formulates aggregation strategy based on the pseudo-labels given by the constructed classifier module. 
Secondly, in the spectral domain, recent work finds the right-shift phenomenon \cite{tang2022rethinking} that the low-frequency component is gradually transferred to the high-frequency part when the degree of fraud becomes more significant. Thus, they try to function like adaptive filters to catch anomalous signals \cite{huang2024cross, ding2022af2gnn, cui2020adaptive}. 
However, due to the significant class imbalance between benign and fraudulent nodes, where benign nodes far outnumber fraudulent ones, high-frequency anomalous signals often account for only a small part, which degenerates the adaptive filter-like GNNs into low-pass filters during training gradually \cite{nt2019revisiting}, thereby failing to avoid the over-smoothing issue. 
Moreover, both spatial and spectral domain methods overlook the increasingly rampant Adaptive Camouflage behaviors of fraudulent users in the real world, for which they struggle to capture anomalous information during the aggregation process. 
However, in our proposed model that focuses on generating healthier relations devoid of Adaptive Camouflage, these weak anomalous signals of Adaptive Camouflage will be enhanced to be obvious enough to be detected instead of being smoothed during the aggregation process

\subsection{Generative Graph Fraud Detection}

Generative models are designed to learn the latent data distribution and generate new samples from this distribution. Variational Autoencoder~(VAE) \cite{kingma2013auto}, Generative Adversarial Network~(GAN) \cite{goodfellow2020generative}, and Denoising Diffusion Probabilistic Model~(DDPM) \cite{ho2020denoising} are three widely used generative models. Due to their remarkable success in various applications of computer vision, they have also been applied to graph-based tasks \cite{yang2023generative, vignac2022digress, meng2023generative}. 
The objective of graph generative models is to generate new graph-level or node-level samples given a set of input graph datasets to augment the original graph\cite{ngo2019fence, zheng2019one, qiao2024generative}, e.g.\ DiGress \cite{vignac2022digress} uses a discrete denoising diffusion model to generate graphs with specific nodes and edge attributes. 
In the context of fraud detection, models like GGA, Fence GAN, and OCAN \cite{meng2023generative, ngo2019fence, zheng2019one} leverage the structure of GAN to directly generate fraudulent nodes located on the benign node boundary, and models like GGAD \cite{qiao2024generative} generates additional anomalous nodes similar to fraudulent nodes by perturbing benign nodes with noise, thus widening the decision boundary between benign and fraudulent nodes. 
Although these methods can utilize generated fraudulent samples to solve common simple camouflage strategies, they still struggle with learning the disguise tactics used in real industry scenarios.
Rather than generating fraudulent samples, which Adaptive Camouflage can mislead, our proposed model concentrates on creating new, homophilic relations. This approach helps to enhance weak anomalous signals and improve the accuracy of final fraud detection.

\section{Methodology}


\subsection{Problem Formulation}
In this paper, we define the original financial networks as graph $G = \{V, R, X \} $, where $V=\{ v_i\ |\ i=1,2,\cdots,n \} $ denotes the set of nodes, $E=\{ R_i\ |\ i=1,2,\cdots,r \}$ denotes the set of relations between nodes, and $X=\{x_i\ |\ i=1,2,\cdots,n \} \in \mathbb{R}^{n\times d} $ denotes the entire feature matrix of nodes, with $d$ denoting the dimension of features.
$|V|=n$ is the number of nodes and $|R|=r$ is the number of relations. 
Every $R_i \in R$ can be represented in the form of an adjacency matrix $\mathcal{A}_i\in \{0,1\}^{n\times n} $.
Furthermore, the set of labels $Y=\{ y_i\ |\ i=1,2,\cdots,n \} $ is defined by experts and denotes whether the node $v_i$ is fraudulent~($y_i=1$) or not~($y_i=0$).

\subsection{Overall Framework}

The overall framework of the model is illustrated in the Figure \ref{fig:model}.
Firstly, the \textbf{Node Group Sampler} module decomposes the original large graph into multiple subgraphs, thereby disrupting the original neighborhood structures where fraudulent nodes have camouflaged connections. This 
effectively dismantle the camouflage behaviors, thus preventing the weak fraudulent signals from being overshadowed by benign information during the aggregation process.
Secondly, we extend contrastive learning methods to a supervised graph training scenario \cite{khosla2020supervised, zhu2021empirical}. 
Our goal is to extract valuable expert knowledge from the limited labels to enhance the difference between fraudulent and benign nodes. Therefore we can leverage it to guide the generation direction to avoid inter-class edges.
Thirdly, the set of decomposed subgraphs is input into the \textbf{Guided Relation Diffusion Generator}. 
During the forward diffusion process, this module iteratively injects a series of normally distributed noises into the subgraphs. 
Then in the subsequent reverse denoising process, the denoising network of Grad learns the true distribution of the original relations by predicting the injected noise. 
During generation processes, the trained \textbf{Supervised Graph Contrastive Learning} module is employed to guide the generation direction based on the similarity between two nodes connected by an edge in the subgraph at timestep $t$ to avoid generating new inter-class edges. 
Simultaneously, a degree penalty function also serves as guidance to ensure that the generated subgraphs adhere closely to the power-law distribution characteristic of financial networks. Based on the above module, subgraphs that are free of inter-class edges and conform to the real financial transaction network distribution are generated, so that the originally weak fraudulent signals would be aggregated iteratively to be obvious enough for the detector module to detect through the aggregation process.
Moreover, the \textbf{Relation Augmentation} module employs a graph diffusion mechanism \cite{gasteiger2019diffusion} based on the Personalized PageRank algorithm to enhance the generated relations by adding extra global information of the entire financial transaction networks to the auxiliary relation. 
Finally, the \textbf{Multi-Relation Detector} captures anomalous information of fraudulent nodes through multiple flexible, spatial/spectral-localized, band-pass filters \cite{tang2022rethinking}. 
The final fraud probability of every node is achieved by using learnable weights on each relation, allowing our proposed model Grad to prioritize the most important and effective relations for different application scenarios.

\subsection{Node Group Sampler}
Fraudulent nodes with Adaptive Camouflage disguise themselves within their local substructures, in which they not only camouflage their node representations but also establish normal connections with benign nodes. 
This would exert a negative impact on the propagation of weak fraudulent signals, so we need to disturb fraudsters' camouflage behaviors.
However, the most common K-hop or Personalized PageRank-based methods for obtaining subgraphs, which most existing works employ, are not useful in this context. 
Because camouflage behaviors in the original graph would be preserved in those subgraph splitting methods, thus weakening the anomalous information of fraudulent nodes during the aggregation process in GNNs.
Additionally, for Grad, retaining these camouflage behaviors adversely affects the quality of the adjacency matrices $\mathcal{A}'$ of subgraphs used for training in the Relation Diffusion Generator module of Section \ref{subsec:DDPM-Guidance}.

Therefore, we design a new subgraph sampling method that begins by randomly shuffling all nodes.
Subsequently, all nodes are divided into multiple equal-sized, non-overlapping groups $\mathcal{G}=\{\mathcal{G}_i \ |\ i=1,2,\dots,\lfloor \frac{n}{k} \rfloor \}$. Each $\mathcal{G}_i $ denotes one node group and $k$ denotes the number of nodes per node group. 
As a result, fraudulent nodes are separated from the original local substructures where they have camouflaged connections. They are instead placed near unfamiliar nodes, which not only diminishes the impact of fraudulent nodes' disguised connections but also assists the Supervised Graph Contrastive Learning module in enhancing the inter-class differences.
Moreover, letting benign nodes interact with other unfamiliar benign nodes and aggregate their information makes the model more robust in identifying benign nodes. The overall process of this module is described in Algorithm \ref{alg:NodeSampler}.

\subsection{Supervised Graph Contrastive Learning}

\label{subsec:SupGCL}
In order to leverage expert knowledge within limited and valuable labels and enhance the difference between the fraudulent and benign nodes, we extend the contrastive learning~(CL) method \cite{khosla2020supervised, zhu2021empirical} to a supervised graph training scenario, in which nodes with the same/different labels are designed to be positive/negative samples of each other.
This supervised graph contrastive learning module can be divided into three parts: the Data Augmentation submodule, the Encoder submodule, and the Project submodule.

In the Data Augmentation submodule, we enhance the node representations by using the graph augment method to obtain an enhanced graph $G_{\text{train}}$,
thus increasing the variety of node representations and improving the module's robustness. 
The enhanced method is computed as $G_{\text{train}}=Aug(G)$,
where $Aug(\cdot)$ is the augment methods such as Edge Adding, Edge Removing, Feature Dropout, Feature Masking, etc \cite{zhu2021empirical}.

In the Encoder submodule, 
to amplify the weak fraudulent signals,
a simple high-pass filter is employed to focus more on the differences between nodes and their neighbors: 
\begin{equation}
    H^{(l+1)}_i=\sigma\ (W^{(l)} \cdot (H^{(l)}_i-Mean(H^{(l)}_{\mathcal{N}_i }))),
\end{equation}
where
$H^{(l+1)}_i$ and $H^{(l)}_i$ are the representation matrices of node $v_i$ in the $(l+1)$-th layer and $(l)$-th layer repectively, $\mathcal{N}_i$ denotes the set of one-hop neighbors of node $v_i$ and $Mean(X_j)$ denotes mean of the neighbors' representations. $\sigma$ is a non-linear activation function, such as ReLU($\cdot$), and $W^{(l)}$ is a weight matrix used to transform the node representations.

Finally, we design the Project submodule to apply the supervised Graph Contrastive Learning module to all nodes $V$ to obtain enhanced representations $X'$. The loss function for a node $v_i$ in this module takes the following form:
\begin{equation}
    \mathcal{L}_{GCL} = \sum_{v_i \in A(i)} \frac{-1}{|{P}(i)|} \sum_{v_p \in {P}(i)} \log \frac{\exp(z_i \cdot z_p / \tau)}{\sum \limits_{v_j \in A(i)} \exp(z_i \cdot z_j / \tau)},
    \label{eq:GCLLoss}
\end{equation}
where, $z_i=\text{Proj}(\text{Encoder}(x_i))\in \mathbb{R}^{h}$, the $\cdot$ denotes the inner product, $\tau \in \mathbb{R}^+$ is a scalar temperature parameter and $h$ denotes the dimensions of node projecting. 
$A(i)=V _{train} \backslash \{v_i\}$ is the set of all training nodes excluding node $ v_i $. $P(i)=\{ v_p\in A(i):y_p=y_i \}$ is the set of all positive samples in batch distinct from $i$, and $|P(i)|$ is its cardinality. This ensures that nodes with the same label have closer representations while those with different labels diverge. 

Moreover, we redesign the module's loss function and calculate its gradient to guide the generation process of the Relation Diffusion Generator module. Based on the validation from directional image generation models \cite{dhariwal2021diffusion}, the distribution of relations predicted at the current step can be guided by the gradient of following formula:
\begin{equation}
\label{eq:Guidance_GCL_sim}
    \text{Gui}_{\text{sim}} = 
    \sum_{v_i \in V} \frac{-1}{|\mathcal{N}_i|} \sum_{v_{\mathcal{N}} \in \mathcal{N}(i)} \log \frac{\exp(z_i \cdot z_\mathcal{N} / \tau)}{\sum \limits_{v_j \in \mathcal{G}^i} \exp(z_i \cdot z_j / \tau)},
\end{equation}
where, 
$\mathcal{G}^i$ denotes the subgraph that contains node $i$. This guidance ensures that in the new relations, nodes tend to connect with others that have more similar node representations, thus reducing inter-class edges during the generation process. 

Through these steps, limited but valuable labels in fraud detection tasks are utilized effectively to widen the decision boundary between benign and fraudulent nodes.
This module is also transformed to guide the sampling process of the Relation Diffusion Generator module. 

\subsection{Guided Relation Diffusion Generator}
\label{subsec:DDPM-Guidance}

    
            
            

            
            
    
    

To make sure that generated relations adhere to the original financial network distribution and avoid inter-class edges, we redesign the running process of the Denoising Diffusion Probabilistic Models~(DDPM) \cite{ho2020denoising, dhariwal2021diffusion, nichol2021improved} as Algorithm \ref{alg:GuidedGeneration}.
In this newly designed training phase, 
we treat the adjacency matrices $\mathcal{A}'$ of node groups $\mathcal{G}$ as image data. 
This module contains two reverse Markov chains: a forward diffusion process and a reverse denoising process. In the forward diffusion process, at each time step $ t $, Gaussian noise $ \epsilon \sim \mathcal{N}(0, \mathbf{I}) $ is injected into the node groups' adjacency matrices $\mathcal{A}'$ according to preset hyperparameter set $\{ \beta^1,\dots,\beta^T \}$:
\begin{equation}
\label{eq:DDPM-foward_one_noise}
    q({\mathcal{A}'}_i^t | {\mathcal{A}'}_i^{t-1}) = \mathcal{N}({\mathcal{A}'}_i^t; \sqrt{1-\beta^t} {\mathcal{A}'}_i^{t-1}, \beta^t \mathbf{I}),
\end{equation}
where ${\mathcal{A}'}_i^t$ denotes the adjacency matrix of $i$-th node group at timestep $t$ and the hyperparameter T denotes the entire timesteps of two diffusion processes. Based on this calculation, the iterative formula of Eq. \ref{eq:DDPM-foward_one_noise} can be computed as: 
\begin{gather}
    \ q({\mathcal{A}'}_i^t | {\mathcal{A}'}_i^0) = \prod_{t=1}^T q({\mathcal{A}'}_i^t | {\mathcal{A}'}_i^{t-1}),\\
    {\mathcal{A}'}_i^t = \sqrt{\bar{\alpha}^t} {\mathcal{A}'}_i^0 + \sqrt{1-\bar{\alpha}^t} \epsilon,
\end{gather}
where $\alpha^t=1-\beta^t$ and $\bar{\alpha}^t=\prod_{s=1}^t \alpha^s$. 
In the reverse denoising process, similar to \cite{ho2020denoising}, a network $\epsilon_{\theta}({\mathcal{A}'}_i^t,t)$ is trained to predict noise. What's more, inspired by directional image generation models from the field of computer vision \cite{ho2020denoising, dhariwal2021diffusion}, we employ an improved graph contrastive learning guidance Eq. \ref{eq:Guidance_GCL_sim} and a degree penalty guidance:
\begin{equation}
\label{eq:Guidance_degree}
    \text{Gui}_{\text{deg}}=
     \sum\nolimits_i \left(\sum\nolimits_j {\mathcal{A}'}_{ij}\right)^2,
\end{equation}
where $ i,j\in [0,k-1]$, to guide the relations generation. At each sampling time step $ t $, the guidance function Eq. \ref{eq:Guidance_GCL_sim} gives guidance by calculating the similarity between currently connected node pairs. The degree penalty function Eq. \ref{eq:Guidance_degree} adjusts the degree distribution of the generated relations, making it more reasonable and more consistent with the degree distribution of financial networks. The entire guidance can be formulated as follows:
\begin{equation}
    \text{Gui}_{\text{all}}=\gamma_1 \cdot \text{Gui}_{\text{sim}}+\gamma_2 \cdot \text{Gui}_{\text{deg}},
\end{equation}
where $\gamma_1$ and $\gamma_2$ are hyperparameters. Then the new noise $\hat{\epsilon}$ takes as follows:
\begin{equation}
    \hat{\epsilon}(\mathcal{A}'^t_i,t) = {\epsilon}_{\theta}(\mathcal{A}'^t_i,t) - s\sqrt{1-\bar{\alpha}^t}\ {\nabla_{\mathcal{A}'^t_i} \log {Gui}_{all}},
\end{equation}
where the $s$ is a hyperparameter to decide the guidance scale.

The setting of variance value $\sigma_t^2$ and the calculation of mean value $\mu_\theta$ are similar to the suggestion in \cite{ho2020denoising}. We use the predicted noise to calculate the mean value $ \mu_\theta $ of adjacency matrices of the node groups based on the partially noised node groups' adjacency matrices, which is formulated as:
\begin{equation}
    \mu_\theta({\mathcal{A}'}_i^t,t) = \frac{1}{\sqrt{\alpha^t}} \left( {\mathcal{A}'}_i^t - \frac{\beta^t}{\sqrt{1-\bar{\alpha}^t}} \hat{\epsilon}(\mathcal{A}'^t_i,t) \right).
\end{equation}

With the variance value $\sigma_t^2$ and the predicted mean value $\mu_\theta$, we can follow the equation:
\begin{equation}
    p_\theta({\mathcal{A}'}_i^{t-1} | {\mathcal{A}'}_i^t) = \mathcal{N}({\mathcal{A}'}_i^{t-1}; \mu_\theta({\mathcal{A}'}_i^t,t), \sigma_t^2 \mathbf{I}),
\end{equation}
to get the sampling of ${\mathcal{A}'}_i^{t-1}$:
\begin{equation}
    {\mathcal{A}'}_i^{t-1} = \mu_\theta({\mathcal{A}'}_i^t,t) + \sigma_t z, \quad \text{with} \quad z \sim \mathcal{N}(0, \mathbf{I}).
\end{equation}

Through these noise addition, noise prediction, and distribution inference training modes, this module directly learns the latent distribution of relations within node groups. 
Therefore we can generate relations that are free of inter-class edges and adhere to the original financial network distribution. 
Based on these homophilic generated relation graphs, the weak fraudulent signals would be enhanced during the aggregation process and become obvious enough for the detection module to detect.

\subsection{Relation Augmentation}

Graph diffusion methods\cite{gasteiger2019diffusion} are employed in this module to ensure that new relations do not overlook global structural information, which contains the trend of the entire transaction networks. Such methods include the Personalized PageRank and the heat kernel, which are applied to the complete network graph formed by merging node groups.
Through these methods, we calculate and establish correlations
between nodes.
Given an adjacency matrix of the entire graph $\mathcal{A} \in \mathbb{R}^{n\times n}$, the graph diffusion is calculated as:
\begin{equation}
    S = \sum_{k=0}^{\infty} \theta_k T^k,
\end{equation}
where $\theta_k$ is overall weighting coefficient among every nodes and $T\in \mathbb{R}^{n\times n}$ is the transition matrix of graph. In our model, we employ the Personalized PageRank method and set $T=\mathcal{A}D^{-1}$. $\theta_k=\varphi(1-\varphi)^k$, where $D\in \mathbb{R}^{n\times n}$ is diagonal degree matrix and $\varphi\in (0,1)$ is the teleport probability. 
To guarantee the convergence of calculation, we set that $\sum_{k=0}^{\infty} \theta_k = 1 \text{, } \theta_k \in [0,1]$, and every eigenvalue $\lambda_i$ of $T$ follows $\lambda_i \in [0,1]$.
Therefore, without multiple steps of iterations, the Personalized PageRank results can be calculated as:
\begin{equation}
    S^{\text{PPR}} = \varphi \left( \mathbf{I}_n - (1 - \varphi) D^{-1/2} \mathcal{A} D^{-1/2} \right)^{-1}.
\end{equation}

\subsection{Multi-Relation Detector}
For each generated auxiliary relation and the original relations, we use $\beta$ kernels as \cite{tang2022rethinking} to capture the anomalous information of fraudulent nodes. 
These kernels contain multiple flexible, spatial/spectral-localized, and band-pass filters which ensure that the important but weak fraudulent signals would be detected and enhanced, instead of being filtered out. More details of $\beta$ kernel are illustrated in Appendix \ref{sec:BetaKernel}. 
For each relation $R_i$, a learnable parameter $\omega_i$ is assigned to determine its weight. This mechanism allows for selecting those most critical relations in various scenarios, thereby obtaining essential node information and enhancing the robustness of our model. The fusion of relations can be defined as:
\begin{equation}
    \text{Grad}(V)=\sum^{r+r'}_{i=1} \omega_i \times f_{\beta}(V,R_i) ,
\end{equation}
where $r'$ denotes the total number of generated auxiliary relations, $f_{\beta}(V,R_i)$ is the output of the $\beta$ wavelet filter on relation $R_i$, and $\text{Grad}(V)$ is the output of entire model.

\subsection{Loss function and Optimization}
For the generation task, as proven in \cite{ho2020denoising}, the denoising network of Grad can be trained by the Mean Square Error Loss~(MSE) between the ground truth of injected noises and the network outputs. So the loss function is:
\begin{equation}
    \mathcal{L}_{\text{Diffusion}}=||{\epsilon}-\hat{\epsilon}(\mathcal{A}'^t_i,t)||.
\end{equation}

Additionally, for the final financial fraud detection task, we treat it as a binary classification task. Given the output of the model $\text{Grad}(V) = \{ \hat{y_i}\ | \ i=1,2,\dots,n \} $ and the ground truth $Y = \{ y_i\ |\ i=1,2,\dots,n \} $, the final financial fraud detection loss is computed as:
\begin{equation}
    \mathcal{L}_{GFD} = - \frac{1}{n} \sum_{i=1}^{n} y_i \log(\hat{y_i}) + (1 - y_i) \log(1 - \hat{y_i}).
\end{equation}

\begin{table*}[ht]
    \centering
    \caption{Performance comparison of different models for fraud detection.}
    \label{tab:main-exp}
    \setlength{\tabcolsep}{8pt}
    \begin{tabular}{l|cc|cc|cc|cc|cc}
        \toprule
        \multirow{2}{*}{Method} & \multicolumn{2}{c|}{Amazon} & \multicolumn{2}{c|}{YelpChi} & \multicolumn{2}{c|}{BlogCatalog} & \multicolumn{2}{c|}{WeChat Pay-Small} & \multicolumn{2}{c}{WeChat Pay-Large}  \\
        \cmidrule(r){2-11}
        & AUC & AP & AUC & AP & AUC & AP & AUC & AP & AUC & AP \\
        \midrule
        Training ratio & \multicolumn{8}{c|}{40\%} & \multicolumn{2}{c}{0.4\%} \\
        \midrule
        MLP                 & 97.04 & 86.95 & 82.42 & 31.27 & 58.15 & 13.30 & 83.59 & 74.55 & 83.45 & 74.44    \\
        SVM                 & 93.76 & 82.80 & 81.34 & 50.25 & 67.13 & 26.44 & 89.59 & 83.18 & 86.51 & 83.19   \\
        GCN                 & 85.32 & 35.14 & 60.34 & 23.69 & 88.09 & 46.52 & 85.24 & 76.83 & 76.02 & 61.22    \\
        GAT                 & 93.04 & 60.67 & 59.82 & 23.48 & 71.25 & 21.45 & 86.11 & 73.58 & 84.88 & 73.91   \\
        GraphSAGE           & 95.85 & 84.74 & 80.44 & 46.83 & 82.34 & 45.09 & 84.24 & 73.92 & 86.21 & 76.20    \\ 
        MixHop              & 96.03 & 86.44 & 79.56 & 45.29 & 88.31 & 48.20 & 90.70 & 84.76 & 86.63 & 84.90    \\
        GPRGNN              & 94.75 & 76.75 & 73.11 & 32.97 & 81.93 & 43.46 & 89.20 & 83.62 & 80.32 & 70.31  \\
        GraphGPS            & 89.35 & 65.02 & \multicolumn{2}{c|}{OOM} & 59.28 & 20.68 & 88.16 & 59.71 & \multicolumn{2}{c}{OOM}  \\
        Graphormer          & 66.81 & 28.97 & \multicolumn{2}{c|}{OOM} & 56.58 & 15.66 & 72.71 & 33.53 & \multicolumn{2}{c}{OOM}  \\
        TGEditor            & 93.65 & 82.66 & 80.43 & 51.35 & 68.51 & 23.61 & 85.39 & 75.45 & \multicolumn{2}{c}{OOM}  \\
        GDC + GCN           & 92.14 & 62.53 & 56.93 & 18.94 & 59.91 & 17.39 & 90.75 & 84.20 & \multicolumn{2}{c}{OOT}  \\
        GGAD                & 94.43 & 79.22 & \multicolumn{2}{c|}{OOM} & 71.35 & 23.64 & 87.16 & 72.85 & \multicolumn{2}{c}{OOM}  \\
        CARE-GNN            & 88.48 & 69.24 & 77.96 & 36.63 & 69.40 & 27.13 & 82.38 & 69.93 & OOM & OOM   \\
        AMNet               & 95.11 & 83.64 & 85.85 & 57.77 & 63.54 & 26.26 & 84.21 & 83.92 & 86.55 & 76.23   \\
        BWGNN               & 97.29 & 88.13 & 90.22 & 63.78 & 79.37 & 37.39 & 90.79 & 84.55 & 86.86 & 76.23   \\
        Grad-w/o Gen        & 97.79 & 88.13 & 84.83 & 54.89 & 76.89 & 29.67 & 90.74 & 84.47 & 77.14 & 63.60   \\
        Grad-w/o Gui        & 96.94 & 87.64 & 92.54 & 73.15 & 77.86 & 29.52 & 90.66 & 84.41 & 81.69 & 68.19   \\
        Grad-w/o WFu        & 91.38 & 60.80 & 66.55 & 29.95 & 93.88 & 67.15 & 88.34 & 88.23 & 86.16 & 75.76   \\
        \textbf{Grad}       & \textbf{97.89} & \textbf{89.68} & \textbf{99.08} & \textbf{96.44} & \textbf{99.41} & \textbf{92.15} & \textbf{91.96} & \textbf{88.17} & \textbf{94.33} & \textbf{90.11} \\
                            & \textbf{+0.60} & \textbf{+1.55} & \textbf{+8.86} & \textbf{+32.66} & \textbf{+11.10} & \textbf{+43.95} & \textbf{+1.17} & \textbf{+3.41} & \textbf{+7.47} & \textbf{+5.21} \\
        \bottomrule
    \end{tabular}
\end{table*}

The proposed model Grad can be optimized through the standard stochastic gradient descent-based algorithms. 
We use Adam optimizer \cite{diederik2014adam} to learn the parameter with $10^{-3}$ learning rate and $10^-5$ weight decay.

\section{Experiments}


\subsection{Experimental Settings}

\subsubsection{Datasets} 

 We collected a large-scale offline industrial dataset from the WeChat Pay platform, provided by Tencent, in compliance with security and privacy policies. This dataset consists of 441,640 users and 110 million related two-hop transactions. 
 The ratio of fraud class is approximately 0.33\%. We set up two different scenarios to utilize this dataset for training and testing the effectiveness of Grad. 
 (1) WeChat Pay-Small Dataset: This dataset follows the training settings of the public dataset and contains only the 4,371 valuable labeled nodes and their 193,137 related transactions. It is divided into training, validation, and testing sets with ratios of 40\%, 30\%, and 30\%, respectively. 
 (2) WeChat Pay-Large Dataset: This dataset is used in a more realistic scenario. 
 Here, only the 4,371 labeled nodes are used for the training and validation process, while the testing is conducted on all the remaining nodes. 
 In this scenario, the training ratio is just 0.4\%, which closely mirrors real-world application scenarios. 
 Moreover, we employed our model in the online industry environment. To prove the versatility of Grad we also conducted experiments in the graph fraud detection scenario using three public datasets: Amazon \cite{mcauley2013amateurs}, YelpChi \cite{rayana2015collective}, and BlogCatalog \cite{tang2009relational}. The statistical information of these datasets is shown in Table \ref{tab:dataAnalysis} in the Appendix.

\subsubsection{Evaluation Metrics}

In our experiments, we use Area Under the ROC Curve~(AUC) and Average Precision~(AP) as evaluation metrics. AUC effectively evaluates binary classification model performance, particularly in the GFD scenario. AP summarizes the precision-recall curve, providing a comprehensive assessment of performance across different threshold levels, which is especially useful for imbalanced datasets. 

\subsubsection{Baselines}

To comprehensively illustrate the effectiveness of Grad, 
we choose several representative models and the SOTA models for comparison. 
More details can be seen in Appendix \ref{sec:AppendixBaselines}.

    
    





\subsubsection{Implementation Details}

For easier reproduction, we list the architectural details of Grad here.
The settings on the node group size $k$, the guidance scale $s$, and the total sample steps $T$ are $32$, $10$, and $100$, respectively. 
Our settings of the Relation Diffusion Generator module and $\beta$ wavelet filter are the same as the setting of Improved DDPM \cite{nichol2021improved} and BWGNN \cite{tang2022rethinking}. 
The hyperparameters of all baselines are consistent with their original papers.

\subsection{Fraud Detection Performance Comparison}

The overall detection performance comparison of Grad and baselines is shown in Table \ref{tab:main-exp}.
Moreover, a case study is described in Appendix~\ref{sec:CaseStudy}.
All the results are averaged over five runs. 
Grad outperforms five benchmarks by different margins, and 
we find:

(1) Grad shows significant improvements on the large dataset, such as the WeChat Pay-Large dataset with 441,640 nodes and the YelpChi dataset with 45,954 nodes. On the WeChat Pay-Large dataset, the AUC increased by 7.47\%, and the AP increased by 5.21\%. On the YelpChi dataset, the AUC index increased by 8.86\%, and the AP index increased by 32.66\%. 
The main reason is that the fraudsters in these two datasets mostly use Adaptive Camouflage to commit fraud.
which decreases the difference in the performance of the fraudsters from the surrounding benign users, narrowing the decision boundary between benign and fraudulent nodes. 
Grad primarily aims to enhance inter-class differences and establish new, reliable relations, rendering fraudsters' disguises within local structures ineffective and making their anomalous activities more conspicuous for the detector.
Furthermore, compared to current Graph Transformer-based and Graph Generative/Augmentation models that are easy to be out of memory~(OOM) when functioning on large datasets, Grad shows a prominent node representing ability and time-memory efficiency for fraud detection.

(2) On small and medium-sized datasets, such as the WeChat Pay-Small dataset with 4371 nodes, the Amazon dataset with 11,944 nodes, and the Blogcatalog dataset with 5,196 nodes, Grad also outperforms existing models. 
The AUC increased by 1.17\%, 0.60\%, and 11.10\%, and the AP increased by 3.41\%, 1.55\%, and 43.95\%, in each dataset respectively. In these small and medium-sized datasets where the characteristics of fraudulent nodes are fairly different from those of benign nodes, the results of each model on the WeChat Pay-Small and Amazon datasets are relatively good. Models have sufficient ability to identify the characteristics of fraudsters. However, in the BlogCatalog data set with 8189-dimensional features, even if salient fraud features exist, the large number of dimensions makes the current model overfit, thus failing to distinguish the characteristics of fraudsters accurately. Our proposed Grad relies on the supervised graph contrastive learning module to reduce the dimensions of features while retaining fraud features. As a result, Grad avoids of being overfitted with meaningless features and captures the fraudulent signals correctly. These observations highlight the effectiveness of the supervised graph contrastive learning module in feature learning and in guiding the construction of new relations without inter-class edges.

\subsection{Ablation Study}

\label{subsec:ablatin_study}


We design three ablation experiments to verify the effectiveness of each module in the model. \textbf{Grad w/o Gen} only uses weighted filters on the original relations in the dataset to capture abnormal information. 
\textbf{Grad w/o Gui} removes the guidance part in the diffusion module to build an unguided model, where the noise predicted by the denoising network is directly used for graph reconstruction. This ablation tests whether the generated graph can construct new relations without camouflage in a purely unsupervised environment. The last ablation, \textbf{Grad w/o WFu} aims to evaluate the performance of the multi-relation fusion wavelet filter in capturing anomalous information. Their performance is compared with the entire Grad framework, as shown in Table \ref{tab:main-exp}.

Overall, the improvement of Grad compared with each ablation verifies our design intention. 
Firstly, the significant declines in the performance of Grad w/o Gen indicate the generation of new undisguised relations has a significant improvement in the model performance under the Adaptive Camouflage. This indirectly proves that the newly generated relations enhance the difference between the fraudsters and the neighbors in behavioral characteristics, making it easier for the detector to capture fraudulent information. 
Secondly, the significant drop in AUC caused by removing guidance verifies that without guidance the generator can not produce auxiliary relations that help the detector to find weak fraudulent signals. 
Thirdly, the results of Grad w/o WFu indicate that the final weighted fusion mechanism can effectively choose the right relations for different scenarios. This effect is more significant in larger datasets like YelpChi and WeChat Pay-Large.

\subsection{Sensitivity Analysis}

\begin{figure}[htbp]
\centering
\includegraphics[width=\linewidth]{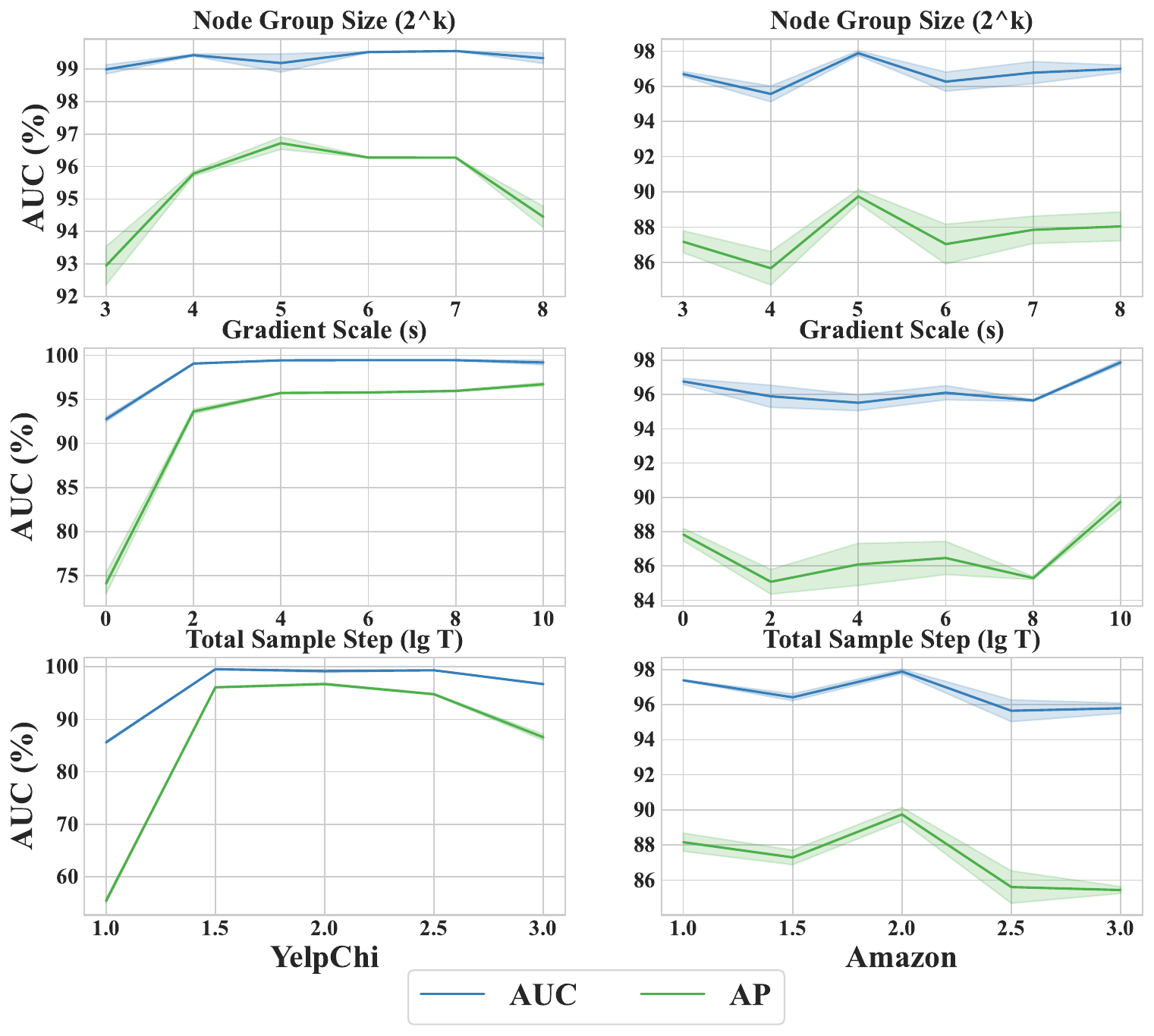}
\caption{ Parameter sensitivity analysis on the node group size $k$, the gradient scale $s$, and the total sample steps $T$. The left sub-figures present the results of experiments conducted on YelpChi, while the right shows the results on Amazon. }

\label{fig:sensitivity}
\end{figure}

In this section, we discuss the important hyperparameters of Grad specifically. The node group size $k$ controls the size of the newly generated local substructure. 
As shown in Figure \ref{fig:sensitivity}, there is a strong correlation between the performance of metrics and the $k$. In general, considering the trade-off between performance and training cost, a node group with a sampling size of 32 is optimal. 
Moreover, on the YelpChi dataset, if the node group size is larger than a certain value, the model performance will deteriorate. 
This is because in the YelpChi dataset, fraudsters have strong disguise capabilities, and the increase in the number of node groups will cause the fraudsters' already weak anomalous information to be smoothed out during the aggregation process. 
We also illustrate the sensitivity analysis of the hyperparameters, sampling time step $T$, and gradient scale $s$ in the guided sampling process. 
First, the YelpChi dataset with a more advanced fraud pattern prefers a larger diffusion step size but at the cost of a longer sampling time. Intuitively, more sampling calculations and guidance are required to generate appropriate new relationships for node groups. 
Secondly, the gradient scale $s$ controls the influence of the guidance function on the sampling process. As the conclusion in the field of image generation, a larger gradient scale can promote more accurate conditional generation [4,37]. However, we found that a larger gradient scale also increases the variance of the model performance metrics AUC and AP. The potential reason is that the drastic modification of the denoising signal may affect the reconstruction of those invisible but regular relations. Furthermore, while enhancing the fraudulent signal, the probability of the fraudulent signal contaminating the benign signal is also increased, which may increase the false positive rate.

\subsection{System Deployment}

\subsubsection{Online Experiments Results}
\begin{table}[htbp]
  \setlength{\tabcolsep}{8pt}
  \caption{Online A/B testing experiments results.}
  \label{tab:onlineExp}
  \begin{tabular}{l|c|c}
    \toprule
    Method & AUC & AP\\
    \midrule
    Running Model-1 & 77.99 & 79.09 \\
    Running Model-2 & 80.30 & 78.81 \\
    Grad & \textbf{93.31} & \textbf{87.96} \\
    \midrule
    improvement & +\textbf{13.01} & +\textbf{8.87} \\
  \bottomrule
\end{tabular}
\end{table}
In this section, an online test on the GFD tasks is conducted on the platform provided by WeChat Pay. 
The objective is to compare the efficacy of our proposed model, Grad, against the graph-based model that has been maturely deployed and operational within the real system. The experimental results are shown in Table \ref{tab:onlineExp}, and AUC and AP are used as the evaluation metrics. Compared to the maturely deployed models, Grad demonstrates remarkable performance improvements across both metrics. Specifically, by effectively mitigating Adaptive Camouflage and capturing important but weak fraudulent information, Grad achieves a \textbf{13.01\%} higher AUC and a significant \textbf{8.87\%} higher AP, which is consistent with the results of the offline comparison in Table \ref{tab:main-exp}. 
This online result also proves that Grad is a robust solution for detecting fraud with disguise.

\subsubsection{Deployment Details}
From the perspective of function, fraud detection consists of in-process detection and post-process detection, which are deployed simultaneously. Every user will be evaluated by the blacklist and fraud rules~(called in-process detection) at first and will be directly blocked if there is a hit on any blacklists or fraud rules. If not, we then extract data from tracking points to generate network relations and integrate relation types and edges to create multiple networks for model input~(Grad here, which is called post-process detection). Subsequently, we load a processed dataset and analyze the outputs from the model. Afterward, the high-risk identified fraudsters are sent for confirmation by domain experts and feedback on the result to the historical database. Finally, we update the model and rules with the latest feedback from experts. During the online implementation, our hardware system has CPUs with AMD EPYC 9K84 and GPUs with NVIDIA A100 SXM4. We conduct experiments in the production environments, using Python 3.8, Pytorch 1.12.1, DGL 0.9.1, and PyG 2.5.2. Regarding costs, Tencent's resource pool leverages a tidal system for optimized GPU usage during peak times, reducing deployment expenses. 

\section{Conclusion}

In this paper, to solve the problem caused by Adaptive Camouflage, in which fraudsters camouflage themselves by mimicking the behavioral data collected by platforms and their key characteristics are highly consistent with those of benign users,
we propose our relation diffusion-based graph augmentation model \textbf{Grad}.
Firstly, we extend a graph contrastive learning module to a supervised graph training scenario to enhance the difference between fraudulent and benign nodes. Then we employ a guided relation diffusion generator to produce homophilic relations to augment weak fraudulent signals. Finally, we use a weighted multi-relation detector to capture weak fraudulent signals through multiple flexible, spatial/spectral-localized, and band-pass filters.
Comprehensive experiments are conducted on online/offline tests on the WeChat Pay platform and three public datasets to prove the effectiveness of Grad,  achieving at most 11.10\% and 43.95\% increases in AUC and AP, respectively. This shows the potential of Grad to protect online payment security and build the trustworthiness of online financial systems.

\begin{acks}
The work is supported by the National Natural Science Foundation of China (62472317), the Fundamental Research Funds for the Central Universities and the Shanghai Science and Technology Innovation Action Plan Project (Grant no. 22YS1400600 and 24692118300).
\end{acks}

\bibliographystyle{www25}
\bibliography{www25}

@String{Computer = "{IEEE} Computer" }

@String{Springer = "Springer-Verlag" }

@article{cheng2023anti,
  title={Anti-money laundering by group-aware deep graph learning},
  author={Cheng, Dawei and Ye, Yujia and Xiang, Sheng and Ma, Zhenwei and Zhang, Ying and Jiang, Changjun},
  journal={IEEE Transactions on Knowledge and Data Engineering},
  volume={35},
  number={12},
  pages={12444--12457},
  year={2023},
  publisher={IEEE}
}

@inproceedings{deng2021graph,
  title={Graph neural network-based anomaly detection in multivariate time series},
  author={Deng, Ailin and Hooi, Bryan},
  booktitle={Proceedings of the AAAI conference on artificial intelligence},
  volume={35},
  number={5},
  pages={4027--4035},
  year={2021}
}

@inproceedings{dou2020enhancing,
  title={Enhancing graph neural network-based fraud detectors against camouflaged fraudsters},
  author={Dou, Yingtong and Liu, Zhiwei and Sun, Li and Deng, Yutong and Peng, Hao and Yu, Philip S},
  booktitle={Proceedings of the 29th ACM international conference on information \& knowledge management},
  pages={315--324},
  year={2020}
}

@article{gao2023rumor,
  title={Rumor detection with self-supervised learning on texts and social graph},
  author={Gao, Yuan and Wang, Xiang and He, Xiangnan and Feng, Huamin and Zhang, Yongdong},
  journal={Frontiers of Computer Science},
  volume={17},
  number={4},
  pages={174611},
  year={2023},
  publisher={Springer}
}

@inproceedings{cheng2021causal,
  title={Causal understanding of fake news dissemination on social media},
  author={Cheng, Lu and Guo, Ruocheng and Shu, Kai and Liu, Huan},
  booktitle={Proceedings of the 27th ACM SIGKDD Conference on Knowledge Discovery \& Data Mining},
  pages={148--157},
  year={2021}
}

@article{liu2024towards,
  title={Towards self-interpretable graph-level anomaly detection},
  author={Liu, Yixin and Ding, Kaize and Lu, Qinghua and Li, Fuyi and Zhang, Leo Yu and Pan, Shirui},
  journal={Advances in Neural Information Processing Systems},
  volume={36},
  year={2024}
}

@inproceedings{ma2023towards,
  title={Towards graph-level anomaly detection via deep evolutionary mapping},
  author={Ma, Xiaoxiao and Wu, Jia and Yang, Jian and Sheng, Quan Z},
  booktitle={Proceedings of the 29th ACM SIGKDD Conference on Knowledge Discovery and Data Mining},
  pages={1631--1642},
  year={2023}
}

@article{qiao2024truncated,
  title={Truncated affinity maximization: One-class homophily modeling for graph anomaly detection},
  author={Qiao, Hezhe and Pang, Guansong},
  journal={Advances in Neural Information Processing Systems},
  volume={36},
  year={2024}
}

@inproceedings{roy2024gad,
  title={Gad-nr: Graph anomaly detection via neighborhood reconstruction},
  author={Roy, Amit and Shu, Juan and Li, Jia and Yang, Carl and Elshocht, Olivier and Smeets, Jeroen and Li, Pan},
  booktitle={Proceedings of the 17th ACM International Conference on Web Search and Data Mining},
  pages={576--585},
  year={2024}
}

@article{zhang2022dual,
  title={Dual-discriminative graph neural network for imbalanced graph-level anomaly detection},
  author={Zhang, Ge and Yang, Zhenyu and Wu, Jia and Yang, Jian and Xue, Shan and Peng, Hao and Su, Jianlin and Zhou, Chuan and Sheng, Quan Z and Akoglu, Leman and others},
  journal={Advances in Neural Information Processing Systems},
  volume={35},
  pages={24144--24157},
  year={2022}
}

@article{yanqiao2020deep,
  title={Deep graph contrastive representation learning},
  author={Yanqiao, Zhu and Yichen, Xu and Feng, Yu and Qiang, Liu and Shu, Wu and Liang, Wang},
  journal={arXiv preprint arXiv:2006.04131},
  year={2020},
  publisher={CoRR}
}

@inproceedings{zhuang2023subgraph,
  title={Subgraph centralization: a necessary step for graph anomaly detection},
  author={Zhuang, Zhong and Ting, Kai Ming and Pang, Guansong and Song, Shuaibin},
  booktitle={Proceedings of the 2023 SIAM International Conference on Data Mining (SDM)},
  pages={703--711},
  year={2023},
  organization={SIAM}
}

@article{mcpherson2001birds,
  title={Birds of a feather: Homophily in social networks},
  author={McPherson, Miller and Smith-Lovin, Lynn and Cook, James M},
  journal={Annual review of sociology},
  volume={27},
  number={1},
  pages={415--444},
  year={2001},
  publisher={Annual Reviews 4139 El Camino Way, PO Box 10139, Palo Alto, CA 94303-0139, USA}
}

@article{zhu2020beyond,
  title={Beyond homophily in graph neural networks: Current limitations and effective designs},
  author={Zhu, Jiong and Yan, Yujun and Zhao, Lingxiao and Heimann, Mark and Akoglu, Leman and Koutra, Danai},
  journal={Advances in neural information processing systems},
  volume={33},
  pages={7793--7804},
  year={2020}
}

@article{hamilton2017inductive,
  title={Inductive representation learning on large graphs},
  author={Hamilton, Will and Ying, Zhitao and Leskovec, Jure},
  journal={Advances in neural information processing systems},
  volume={30},
  year={2017}
}

@article{kipf2016semi,
  title={Semi-supervised classification with graph convolutional networks},
  author={Kipf, Thomas N and Welling, Max},
  journal={arXiv preprint arXiv:1609.02907},
  year={2016}
}

@article{velivckovic2017graph,
  title={Graph attention networks},
  author={Veli{\v{c}}kovi{\'c}, Petar and Cucurull, Guillem and Casanova, Arantxa and Romero, Adriana and Lio, Pietro and Bengio, Yoshua},
  journal={arXiv preprint arXiv:1710.10903},
  year={2017}
}

@article{nt2019revisiting,
  title={Revisiting graph neural networks: All we have is low-pass filters},
  author={Nt, Hoang and Maehara, Takanori},
  journal={arXiv preprint arXiv:1905.09550},
  year={2019}
}

@article{zhang2024generation,
  title={Generation is better than Modification: Combating High Class Homophily Variance in Graph Anomaly Detection},
  author={Zhang, Rui and Cheng, Dawei and Liu, Xin and Yang, Jie and Ouyang, Yi and Wu, Xian and Zheng, Yefeng},
  journal={arXiv preprint arXiv:2403.10339},
  year={2024}
}

@inproceedings{meng2023generative,
  title={Generative Graph Augmentation for Minority Class in Fraud Detection},
  author={Meng, Lin and Mostafa, Hesham and Nassar, Marcel and Zhang, Xiaonan and Zhang, Jiawei},
  booktitle={Proceedings of the 32nd ACM International Conference on Information and Knowledge Management},
  pages={4200--4204},
  year={2023}
}

@article{qiao2024generative,
  title={Generative Semi-supervised Graph Anomaly Detection},
  author={Qiao, Hezhe and Wen, Qingsong and Li, Xiaoli and Lim, Ee-Peng and Pang, Guansong},
  journal={arXiv preprint arXiv:2402.11887},
  year={2024}
}

@inproceedings{wu2023graph,
  title={Graph contrastive learning with generative adversarial network},
  author={Wu, Cheng and Wang, Chaokun and Xu, Jingcao and Liu, Ziyang and Zheng, Kai and Wang, Xiaowei and Song, Yang and Gai, Kun},
  booktitle={Proceedings of the 29th ACM SIGKDD Conference on Knowledge Discovery and Data Mining},
  pages={2721--2730},
  year={2023}
}

@inproceedings{liu2021pick,
  title={Pick and choose: a GNN-based imbalanced learning approach for fraud detection},
  author={Liu, Yang and Ao, Xiang and Qin, Zidi and Chi, Jianfeng and Feng, Jinghua and Yang, Hao and He, Qing},
  booktitle={Proceedings of the web conference 2021},
  pages={3168--3177},
  year={2021}
}

@inproceedings{tang2022rethinking,
  title={Rethinking graph neural networks for anomaly detection},
  author={Tang, Jianheng and Li, Jiajin and Gao, Ziqi and Li, Jia},
  booktitle={International Conference on Machine Learning},
  pages={21076--21089},
  year={2022},
  organization={PMLR}
}

@inproceedings{gao2023addressing,
  title={Addressing heterophily in graph anomaly detection: A perspective of graph spectrum},
  author={Gao, Yuan and Wang, Xiang and He, Xiangnan and Liu, Zhenguang and Feng, Huamin and Zhang, Yongdong},
  booktitle={Proceedings of the ACM Web Conference 2023},
  pages={1528--1538},
  year={2023}
}

@inproceedings{ghosh2023gosage,
  title={GoSage: Heterogeneous Graph Neural Network Using Hierarchical Attention for Collusion Fraud Detection},
  author={Ghosh, Soumava and Anand, Ravi and Bhowmik, Tanmoy and Chandrashekhar, Siddhanth},
  booktitle={Proceedings of the Fourth ACM International Conference on AI in Finance},
  pages={185--192},
  year={2023}
}

@article{tian2023sad,
  title={Sad: Semi-supervised anomaly detection on dynamic graphs},
  author={Tian, Sheng and Dong, Jihai and Li, Jintang and Zhao, Wenlong and Xu, Xiaolong and Song, Bowen and Meng, Changhua and Zhang, Tianyi and Chen, Liang and others},
  journal={arXiv preprint arXiv:2305.13573},
  year={2023}
}

@inproceedings{duan2024dga,
  title={DGA-GNN: Dynamic Grouping Aggregation GNN for Fraud Detection},
  author={Duan, Mingjiang and Zheng, Tongya and Gao, Yang and Wang, Gang and Feng, Zunlei and Wang, Xinyu},
  booktitle={Proceedings of the AAAI Conference on Artificial Intelligence},
  volume={38},
  number={10},
  pages={11820--11828},
  year={2024}
}

@inproceedings{wang2019semi,
  title={A semi-supervised graph attentive network for financial fraud detection},
  author={Wang, Daixin and Lin, Jianbin and Cui, Peng and Jia, Quanhui and Wang, Zhen and Fang, Yanming and Yu, Quan and Zhou, Jun and Yang, Shuang and Qi, Yuan},
  booktitle={2019 IEEE international conference on data mining (ICDM)},
  pages={598--607},
  year={2019},
  organization={IEEE}
}

@inproceedings{cui2020deterrent,
  title={Deterrent: Knowledge guided graph attention network for detecting healthcare misinformation},
  author={Cui, Limeng and Seo, Haeseung and Tabar, Maryam and Ma, Fenglong and Wang, Suhang and Lee, Dongwon},
  booktitle={Proceedings of the 26th ACM SIGKDD international conference on knowledge discovery \& data mining},
  pages={492--502},
  year={2020}
}

@article{ho2020denoising,
  title={Denoising diffusion probabilistic models},
  author={Ho, Jonathan and Jain, Ajay and Abbeel, Pieter},
  journal={Advances in neural information processing systems},
  volume={33},
  pages={6840--6851},
  year={2020}
}

@article{dhariwal2021diffusion,
  title={Diffusion models beat gans on image synthesis},
  author={Dhariwal, Prafulla and Nichol, Alexander},
  journal={Advances in neural information processing systems},
  volume={34},
  pages={8780--8794},
  year={2021}
}

@article{khosla2020supervised,
  title={Supervised contrastive learning},
  author={Khosla, Prannay and Teterwak, Piotr and Wang, Chen and Sarna, Aaron and Tian, Yonglong and Isola, Phillip and Maschinot, Aaron and Liu, Ce and Krishnan, Dilip},
  journal={Advances in neural information processing systems},
  volume={33},
  pages={18661--18673},
  year={2020}
}

@article{zhu2021empirical,
  title={An empirical study of graph contrastive learning},
  author={Zhu, Yanqiao and Xu, Yichen and Liu, Qiang and Wu, Shu},
  journal={arXiv preprint arXiv:2109.01116},
  year={2021}
}

@article{gasteiger2019diffusion,
  title={Diffusion improves graph learning},
  author={Gasteiger, Johannes and Wei{\ss}enberger, Stefan and G{\"u}nnemann, Stephan},
  journal={Advances in neural information processing systems},
  volume={32},
  year={2019}
}

@article{goodfellow2020generative,
  title={Generative adversarial networks},
  author={Goodfellow, Ian and Pouget-Abadie, Jean and Mirza, Mehdi and Xu, Bing and Warde-Farley, David and Ozair, Sherjil and Courville, Aaron and Bengio, Yoshua},
  journal={Communications of the ACM},
  volume={63},
  number={11},
  pages={139--144},
  year={2020},
  publisher={ACM New York, NY, USA}
}

@inproceedings{zheng2019one,
  title={One-class adversarial nets for fraud detection},
  author={Zheng, Panpan and Yuan, Shuhan and Wu, Xintao and Li, Jun and Lu, Aidong},
  booktitle={Proceedings of the AAAI Conference on Artificial Intelligence},
  volume={33},
  number={01},
  pages={1286--1293},
  year={2019}
}

@inproceedings{ngo2019fence,
  title={Fence GAN: Towards better anomaly detection},
  author={Ngo, Phuc Cuong and Winarto, Amadeus Aristo and Kou, Connie Khor Li and Park, Sojeong and Akram, Farhan and Lee, Hwee Kuan},
  booktitle={2019 IEEE 31St International Conference on tools with artificial intelligence (ICTAI)},
  pages={141--148},
  year={2019},
  organization={IEEE}
}

@article{yang2024directional,
  title={Directional diffusion models for graph representation learning},
  author={Yang, Run and Yang, Yuling and Zhou, Fan and Sun, Qiang},
  journal={Advances in Neural Information Processing Systems},
  volume={36},
  year={2024}
}

@article{kingma2013auto,
  title={Auto-encoding variational bayes},
  author={Kingma, Diederik P and Welling, Max},
  journal={arXiv preprint arXiv:1312.6114},
  year={2013}
}

@article{vignac2022digress,
  title={Digress: Discrete denoising diffusion for graph generation},
  author={Vignac, Clement and Krawczuk, Igor and Siraudin, Antoine and Wang, Bohan and Cevher, Volkan and Frossard, Pascal},
  journal={arXiv preprint arXiv:2209.14734},
  year={2022}
}

@inproceedings{yang2023generative,
  title={Generative-contrastive graph learning for recommendation},
  author={Yang, Yonghui and Wu, Zhengwei and Wu, Le and Zhang, Kun and Hong, Richang and Zhang, Zhiqiang and Zhou, Jun and Wang, Meng},
  booktitle={Proceedings of the 46th International ACM SIGIR Conference on Research and Development in Information Retrieval},
  pages={1117--1126},
  year={2023}
}

@inproceedings{huang2024cross,
  title={Cross-Space Adaptive Filter: Integrating Graph Topology and Node Attributes for Alleviating the Over-smoothing Problem},
  author={Huang, Chen and Li, Haoyang and Zhang, Yifan and Lei, Wenqiang and Lv, Jiancheng},
  booktitle={Proceedings of the ACM on Web Conference 2024},
  pages={803--814},
  year={2024}
}

@article{ding2022af2gnn,
  title={AF2GNN: Graph convolution with adaptive filters and aggregator fusion for hyperspectral image classification},
  author={Ding, Yao and Zhang, Zhili and Zhao, Xiaofeng and Hong, Danfeng and Li, Wei and Cai, Wei and Zhan, Ying},
  journal={Information Sciences},
  volume={602},
  pages={201--219},
  year={2022},
  publisher={Elsevier}
}

@inproceedings{cui2020adaptive,
  title={Adaptive graph encoder for attributed graph embedding},
  author={Cui, Ganqu and Zhou, Jie and Yang, Cheng and Liu, Zhiyuan},
  booktitle={Proceedings of the 26th ACM SIGKDD international conference on knowledge discovery \& data mining},
  pages={976--985},
  year={2020}
}

@inproceedings{mcauley2013amateurs,
  title={From amateurs to connoisseurs: modeling the evolution of user expertise through online reviews},
  author={McAuley, Julian John and Leskovec, Jure},
  booktitle={Proceedings of the 22nd international conference on World Wide Web},
  pages={897--908},
  year={2013}
}

@inproceedings{rayana2015collective,
  title={Collective opinion spam detection: Bridging review networks and metadata},
  author={Rayana, Shebuti and Akoglu, Leman},
  booktitle={Proceedings of the 21th acm sigkdd international conference on knowledge discovery and data mining},
  pages={985--994},
  year={2015}
}

@inproceedings{tang2009relational,
  title={Relational learning via latent social dimensions},
  author={Tang, Lei and Liu, Huan},
  booktitle={Proceedings of the 15th ACM SIGKDD international conference on Knowledge discovery and data mining},
  pages={817--826},
  year={2009}
}

@article{chien2020adaptive,
  title={Adaptive universal generalized pagerank graph neural network},
  author={Chien, Eli and Peng, Jianhao and Li, Pan and Milenkovic, Olgica},
  journal={arXiv preprint arXiv:2006.07988},
  year={2020}
}

@inproceedings{abu2019mixhop,
  title={Mixhop: Higher-order graph convolutional architectures via sparsified neighborhood mixing},
  author={Abu-El-Haija, Sami and Perozzi, Bryan and Kapoor, Amol and Alipourfard, Nazanin and Lerman, Kristina and Harutyunyan, Hrayr and Ver Steeg, Greg and Galstyan, Aram},
  booktitle={international conference on machine learning},
  pages={21--29},
  year={2019},
  organization={PMLR}
}

@inproceedings{boser1992training,
  title={A training algorithm for optimal margin classifiers},
  author={Boser, Bernhard E and Guyon, Isabelle M and Vapnik, Vladimir N},
  booktitle={Proceedings of the fifth annual workshop on Computational learning theory},
  pages={144--152},
  year={1992}
}

@article{rosenblatt1958perceptron,
  title={The perceptron: a probabilistic model for information storage and organization in the brain.},
  author={Rosenblatt, Frank},
  journal={Psychological review},
  volume={65},
  number={6},
  pages={386},
  year={1958},
  publisher={American Psychological Association}
}

@inproceedings{chai2022can,
  title={Can abnormality be detected by graph neural networks?},
  author={Chai, Ziwei and You, Siqi and Yang, Yang and Pu, Shiliang and Xu, Jiarong and Cai, Haoyang and Jiang, Weihao},
  booktitle={IJCAI},
  pages={1945--1951},
  year={2022}
}

@inproceedings{nichol2021improved,
  title={Improved denoising diffusion probabilistic models},
  author={Nichol, Alexander Quinn and Dhariwal, Prafulla},
  booktitle={International conference on machine learning},
  pages={8162--8171},
  year={2021},
  organization={PMLR}
}

@techreport{commission2023consumer,
    author = {Federal Trade Commission},
    title = {Consumer Sentinel Network Data Book 2023},
    institution = {Federal Trade Commission},
    year = {2023}
}

@inproceedings{zhang2024pre,
  title={Pre-trained Online Contrastive Learning for Insurance Fraud Detection},
  author={Zhang, Rui and Cheng, Dawei and Yang, Jie and Ouyang, Yi and Wu, Xian and Zheng, Yefeng and Jiang, Changjun},
  booktitle={Proceedings of the AAAI Conference on Artificial Intelligence},
  volume={38},
  number={20},
  pages={22511--22519},
  year={2024}
}

@article{rampavsek2022recipe,
  title={Recipe for a general, powerful, scalable graph transformer},
  author={Ramp{\'a}{\v{s}}ek, Ladislav and Galkin, Michael and Dwivedi, Vijay Prakash and Luu, Anh Tuan and Wolf, Guy and Beaini, Dominique},
  journal={Advances in Neural Information Processing Systems},
  volume={35},
  pages={14501--14515},
  year={2022}
}

@article{ying2021transformers,
  title={Do transformers really perform badly for graph representation?},
  author={Ying, Chengxuan and Cai, Tianle and Luo, Shengjie and Zheng, Shuxin and Ke, Guolin and He, Di and Shen, Yanming and Liu, Tie-Yan},
  journal={Advances in neural information processing systems},
  volume={34},
  pages={28877--28888},
  year={2021}
}

@inproceedings{zhang2023tgeditor,
  title={TGEditor: Task-Guided Graph Editing for Augmenting Temporal Financial Transaction Networks},
  author={Zhang, Shuaicheng and Zhu, Yada and Zhou, Dawei},
  booktitle={Proceedings of the Fourth ACM International Conference on AI in Finance},
  pages={219--226},
  year={2023}
}

@article{khedmati2020applying,
  title={Applying support vector data description for fraud detection},
  author={Khedmati, Mohamad and Erfani, Masoud and GhasemiGol, Mohammad},
  journal={arXiv preprint arXiv:2006.00618},
  year={2020}
}

@inproceedings{wu2024safeguarding,
  title={Safeguarding Fraud Detection from Attacks: A Robust Graph Learning Approach},
  author={Wu, Jiasheng and Liu, Xin and Cheng, Dawei and Ouyang, Yi and Wu, Xian and Zheng, Yefeng},
  booktitle={Proceedings of the Thirty-Third International Joint Conference on Artificial Intelligence, IJCAI-24},
  volume={8},
  pages={7500--7508},
  year={2024}
}

@article{diederik2014adam,
  title={Adam: A method for stochastic optimization},
  author={Diederik, P Kingma},
  journal={(No Title)},
  year={2014}
}

@article{cheng2022regulating,
  title={Regulating systemic crises: Stemming the contagion risk in networked-loans through deep graph learning},
  author={Cheng, Dawei and Niu, Zhibin and Li, Jie and Jiang, Changjun},
  journal={IEEE Transactions on Knowledge and Data Engineering},
  volume={35},
  number={6},
  pages={6278--6289},
  year={2022},
  publisher={IEEE}
}

@article{cheng2018modeling,
  title={Modeling similarities among multi-dimensional financial time series},
  author={Cheng, Dawei and Liu, Ye and Niu, Zhibin and Zhang, Liqing},
  journal={IEEE Access},
  volume={6},
  pages={43404--43413},
  year={2018},
  publisher={IEEE}
}

@inproceedings{ma2023fighting,
  title={Fighting against Organized Fraudsters Using Risk Diffusion-based Parallel Graph Neural Network.},
  author={Ma, Jiacheng and Li, Fan and Zhang, Rui and Xu, Zhikang and Cheng, Dawei and Ouyang, Yi and Zhao, Ruihui and Zheng, Jianguang and Zheng, Yefeng and Jiang, Changjun},
  booktitle={IJCAI},
  pages={6138--6146},
  year={2023}
}

@article{cheng2025graph,
  title={Graph neural networks for financial fraud detection: a review},
  author={Cheng, Dawei and Zou, Yao and Xiang, Sheng and Jiang, Changjun},
  journal={Frontiers of Computer Science},
  volume={19},
  number={9},
  pages={1--15},
  year={2025},
  publisher={Springer}
}

@article{han2025mitigating,
  title={Mitigating the Tail Effect in Fraud Detection by Community Enhanced Multi-Relation Graph Neural Networks},
  author={Han, Li and Wang, Longxun and Cheng, Ziyang and Wang, Bo and Yang, Guang and Cheng, Dawei and Lin, Xuemin},
  journal={IEEE Transactions on Knowledge and Data Engineering},
  year={2025},
  publisher={IEEE}
}

\appendix

\begin{table*}[htbp]
\centering
\caption{Statistics of four anomaly detection datasets.}
\label{tab:dataAnalysis}
\newcolumntype{C}{>{\centering\arraybackslash}X} 
\newcolumntype{Y}[1]{>{\centering\arraybackslash}p{#1}}
\begin{tabularx}{0.95\textwidth}{@{}Y{2cm}|CY{2.5cm}CCY{1.5cm}CCC@{}}
\toprule
\textbf{Dataset} & \textbf{Type}& \textbf{Scenarios} & \textbf{Node} & \textbf{Relations} & \textbf{Edge} &\textbf{Features} & \textbf{Anomalies} & \textbf{Fraud Rate} \\
\midrule
\multirow{3}{*}{YelpChi}&\multirow{3}{*}{Real} & \multirow{3}{*}{Review} & \multirow{3}{*}{45,954} & R-U-R & 98.630   &\multirow{3}{*}{32}& \multirow{3}{*}{6,674} & \multirow{3}{*}{14.52\%} \\
                       &  &                         &                        & R-S-R & 6,805,486 &&                    & \\
                       &  &                         &                        & R-T-R & 1,147,232   &&                    & \\
\midrule
\multirow{3}{*}{Amazon}&\multirow{3}{*}{Real} & \multirow{3}{*}{Review} & \multirow{3}{*}{11,944} & U-P-U & 351,216   &\multirow{3}{*}{25}& \multirow{3}{*}{821} & \multirow{3}{*}{9.50\%} \\
                       & &                         &                        & U-S-U & 7,132,958  &&                    & \\
                       & &                         &                         & U-V-U & 2,073,474 &&                    & \\
\midrule
BlogCatalog & Inject &Social Networks & 5,196   & - & 171,743 &8,189& 300 & 5.77\% \\
\midrule
WeChat Pay-Small & Real &Social Networks & 4,371   & - & 193,137 &50& 426 & 9.75\% \\
\midrule
WeChat Pay-Large & Real &Social Networks & 441,640   & - & 2,196,022 &50& 1464 & 0.33\% \\
\bottomrule
\end{tabularx}
\end{table*}

\section{Details of $\beta$ Kernels}
\label{sec:BetaKernel}
The $\beta$ kernel is set as follows:
\begin{equation}
    \beta_{p,q}(w)=
        \begin{cases}
        \frac{1}{B(p+1,q+1)}w^p(1-w)^q&\text{if }w\in[0,1]\\
        0&\text{otherwise}
        \end{cases}
\end{equation}
Here, $p,q\in\mathbb{R}^+$. To ensure $w\in[0,2]$ meets the eigenvalue condition for normalized Laplacian matrix, and to express $\beta_{p,q}(w)$ in polynomial form for faster computing, we have
\begin{equation}
    \beta^*_{p,q}(w)=\frac{1}{2}\beta_{p,q}\left(\frac{w}{2}\right),
\end{equation}
with $p,q\in\mathbb{N}^+$. The final kernel is:
\begin{equation}
    \mathcal{W}_{p,q}=U\beta^*_{p,q}(\Lambda)U^T=\beta^*_{p,q}(L)=\frac{\left(\frac{L}{2}\right)^p\left(I-\frac{L}{2}\right)^q}{2B(p+1,q+1)}.
\end{equation}

\begin{algorithm} [htbp]
\caption{Guided Relation Diffusion Generation}
\label{alg:GuidedGeneration}
    \begin{algorithmic}[1]
    \Require Sampled Node Group $\mathcal{G}_i$, Total Sample Step $T$, Denoising network $\epsilon_\theta$, Variance schedule $\{\beta^t \}$, and Gradient scale $s$
    
        \For{all $t$ from $T$ to $1$}
            
            \State ${Gui}_{all}^{i} \gets \gamma_1 \cdot {Gui}_{sim}^{i}+\gamma_2 \cdot {Gui}_{deg}^{i}$
            
            \State $\hat{\epsilon}(\mathcal{A}'^t_i,t) \gets {\epsilon}_{\theta}(\mathcal{A}'^t_i,t) - s\sqrt{1-\bar{\alpha}^t} \nabla_{\mathcal{A}'^t_i} \log {Gui}_{all}^{i}$

            
            \State ${\mathcal{A}'}_i^{t-1} \gets \frac{1}{\sqrt{\alpha^t}} \left( {\mathcal{A}'}_i^t - \frac{\beta^t}{\sqrt{1-\bar{\alpha}^t}} \hat{\epsilon}(\mathcal{A}'^t_i,t) \right) + \sqrt{\beta^t} z $
            
        \EndFor
    
    \State \textbf{return} $\mathcal{G}_i$
    
    \end{algorithmic}
\end{algorithm}

\begin{algorithm}
    \caption{{Entire Algorithm of Grad}}
    \begin{algorithmic}[1]
        \State \textbf{Input:} Nodes $V$, Number of Nodes $n$, Group Size $k$, Raw Relations $R$, Total Sample Step $T$, Variance Schedule $\beta^T$, Gradient Scale $s$
        \State \textbf{Output:} Detection Result $P$

        \State $\mathcal{R} \gets$ Fuse Relations(R) 

        \State $\mathcal{G}$ $\gets$ Node Group Sampling($V$, $n$, $k$)

        \State Train Supervised Graph Contrastive Learning Module($V$, $\mathcal{G}$, $\mathcal{R}$)

        \State $\mathtt{R}$ $\gets$ Guided Relation Generation($\mathcal{G}$, $T$, $\{ \beta^t \}$, $s$)

        \State $\mathtt{R}'$ $\gets$ Graph Diffusion Convolution($\mathtt{R}$)

        \State $P$ $\gets$ Weighted Wavelet Filter($V$, $\mathtt{R}'$)
        
        \State \textbf{Return} $P$
    \end{algorithmic}
\end{algorithm}

\begin{algorithm}
    \caption{{Node Group Sampling Algorithm}}
    \label{alg:NodeSampler}
    \begin{algorithmic}[1]
        \State \textbf{Input:} Nodes $V$, Number of Nodes $n$, Group Size $k$
        \State \textbf{Output:} List of Node Groups $\mathcal{G}$
        
        \State $V \gets$ Shuffle($V$)
        
        \For{$i \gets 1$ to $n$}
            \State $V[i].index \gets i$
        \EndFor
        
        \State $group\_size \gets k$
        \State $group\_num \gets \lfloor \frac{n}{k} \rfloor$
        \State $\mathcal{G} \gets$ Empty list of groups
        
        \For{$i \gets 0$ to $group\_num-1$}
            \State $\mathcal{G}_i \gets V[i \times group\_size : (i+1) \times group\_size]$
            \State Append $\mathcal{G}_i$ to $\mathcal{G}$
        \EndFor
        
        \State \textbf{Return} $\mathcal{G}$
    \end{algorithmic}
\end{algorithm}



\section{Baselines}
\label{sec:AppendixBaselines}

\begin{enumerate}[1.]
    \item Classic machine learning models
        \begin{itemize}
            \item \textbf{MLP} \cite{rosenblatt1958perceptron}: it is a type of feedforward neural network with one or more hidden layers between the input and output layers. 
            
            \item \textbf{SVM} \cite{boser1992training}: It is a supervised learning model used for classification and regression analysis that finds an optimal hyperplane that maximally separates different classes.
        \end{itemize}
    
    \item Basic graph-based models
        \begin{itemize}
            \item \textbf{GCN} \cite{kipf2016semi}: It is a neural network architecture that applies convolutional operations on graph-structured data to learn node embeddings and make predictions based on the graph's topology and features.
            
            \item \textbf{GAT} \cite{velivckovic2017graph}: It is a neural network architecture that uses masked self-attention mechanisms to weigh the importance of neighboring nodes in a graph when learning node representations.
            
            \item \textbf{GraphSAGE} \cite{hamilton2017inductive}: It is a neural network architecture that employs an inductive framework to efficiently generate node embeddings by aggregating and propagating feature information from a node's local neighborhood in a graph.
        \end{itemize}
        
    \item heterophilic graph models
        \begin{itemize}
            \item \textbf{MixHop} \cite{abu2019mixhop}: It utilizes a novel graph convolutional layer structure that improves feature learning on graph data by blending information from neighbors at different hops.
            
            \item \textbf{GPRGNN} \cite{chien2020adaptive}: It combines the general PageRank algorithm with an adaptive mechanism for node importance assessment and attribute prediction on graph data.
        \end{itemize}

    \item Graph Transformer-based models
        \begin{itemize}
            \item \textbf{GraphGPS} \cite{rampavsek2022recipe}: It is a general, powerful, and scalable Graph Transformer with linear complexity, which supports multiple types of encoding methods and is applicable to graph structures of different sizes.
            
            \item \textbf{Graphormer} \cite{ying2021transformers}: It utilizes the powerful sequence modeling capability of Transformer, a self-attention structure with a global receptive field, and introduces three spatial encoding methods to make up for the lack of Transformer's ability to perceive graph structures.
            
        \end{itemize}

    \item Graph Transformer-based models
        \begin{itemize}
            \item \textbf{TGEditor} \cite{zhang2023tgeditor}: It preserves the temporal and topological distribution of input financial transaction networks, whilst leveraging the label information from pertinent downstream tasks inclusive of crucial downstream tasks like fraudulent transaction classification. 
            
            \item \textbf{GDC} \cite{gasteiger2019diffusion}: It employs graph diffusion methods, including Personal PageRank and Heat Kernel, to reconstruct the original graph relation.

            \item \textbf{GGAD} \cite{qiao2024generative}: It generates additional anomalous nodes similar to fraudulent nodes by perturbing benign nodes with noise to solve the class imbalance and improve model's robustness.
            
        \end{itemize}
    
    \item GAD-based models
        \begin{itemize}
            \item \textbf{CARE-GNN} \cite{dou2020enhancing}: It utilizes label-aware similarity to identify neighborhoods, employs reinforcement learning to determine the optimal number of neighbors, and aggregates selected neighbors across different relationships for enhanced graph-based learning.
            
            \item \textbf{AMNet} \cite{chai2022can}: It adaptively combines multi-frequency signals to improve anomaly detection in graphs by capturing both local and global structural anomalies.
            
            \item \textbf{BWGNN} \cite{tang2022rethinking}: It leverages spectral and spatial localized bandpass filters to enhance anomaly detection in graphs, effectively addressing the 'right-shift' spectral phenomenon and improving the identification of anomalies.
        \end{itemize}
\end{enumerate}

\section{Case Study}
\label{sec:CaseStudy}

\begin{figure}[htbp]
\centering
\includegraphics[width=\linewidth]{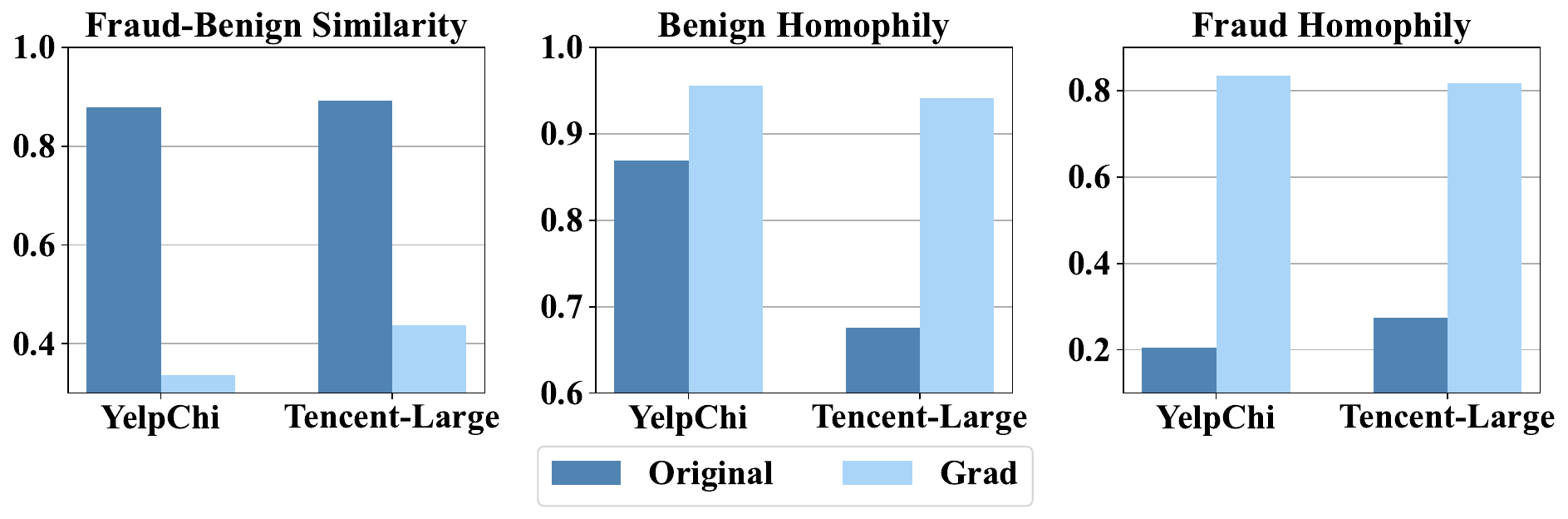}
\caption{ Comparation on original and Grad-generated YelpChi and WeChat Pay-Large datasets. 
}

\label{fig:gradVSori}
\end{figure}

We visualize some properties of the original and Grad-generated graph in Figure \ref{fig:gradVSori}. The original seriously high fraud-benign similarity ratios, caused by Adaptive Camouflage, are detected and eased by the Supervised Graph Contrastive Learning module proposed in Section \ref{subsec:SupGCL}. The significant drops in the YelpChi and the WeChat Pay-Large dataset shown in the left part (from over 80\% to around 30\%) provide the model easy access to identifying the important anomalous information. What's more, as shown in the middle and right part, homophily ratios of both benign nodes and fraudulent nodes in the Grad-generated graphs have huge improvements.
The homophily ratios of benign nodes increased from 62.9\% and \% 54.2\% to over 90\% in the YelpChi and WeChat Pay-Large datasets respectively. 
For fraudulent nodes, the homophily ratios are improved from 8.70 and 26.5 to over 80\% respectively.
As a result, the newly generated graphs are more homophilic, thus meeting with the homophily assumption of current GNNs \cite{mcpherson2001birds, zhu2020beyond, nt2019revisiting}, which enables GNNs to amplify the weak fraudulent signals through aggregation process.

\end{document}